# On the limits of algorithmic prediction across the globe

Xingyu Li[1], Difan Song[2], Miaozhe Han[3], Yu Zhang[4], and René F. Kizilcec[5]

[1] Stanford University, Stanford, CA, USA, 94305

[2] Shanghai Advanced Institute of Finance, Shanghai Jiao Tong University, Shanghai, China, 200030

[3] The Chinese University of Hong Kong, Hong Kong SAR, China, 999077

[4] Fudan University, Shanghai, China, 200433

[5] Cornell University, Ithaca, NY, USA, 14850

**Abstract**

The impact of predictive algorithms on people's lives and livelihoods has been noted in medicine, criminal justice, finance, hiring and admissions. Most of these algorithms are developed using data and human capital from highly developed nations. We tested how well predictive models of human behavior trained in a developed country generalize to people in less developed countries by modeling global variation in 200 predictors of academic achievement on nationally representative student data for 65 countries. Here we show that state-of-the-art machine learning models trained on data from the United States can predict achievement with high accuracy and generalize to other developed countries with comparable accuracy. However, accuracy drops linearly with national development due to global variation in the importance of different achievement predictors, providing a useful heuristic for policymakers. Training the same model on national data yields high accuracy in every country, which highlights the value of local data collection.

**Significance Statement**

Ninety-six percent of datasets in the behavioral sciences and 80% in education originate from a single country–the United States. This raises an important question: how well do models of human behavior developed in the U.S. generalize to other people, especially those in less developed countries? The current research provides the first piece of empirical evidence and showed that a U.S.-trained model of academic achievement performs no better than chance in some parts of the developing world. This is because the factors that motivate human behavior and performance vary substantially across the globe, and thus what matters in predicting human behavior in the United States (such as books and personal aspiration) is different to what matters in the developing world.

**Main Text**

People's lives around the world are increasingly affected by algorithms and big data (1, 2). Algorithms are used to predict personal preferences and behaviors to allocate resources and recommend actions in countless domains. They assist medical professionals in diagnosing cancer (3), judges in deciding which defendants awaiting trial to release on bail (4, 5), banks in offering loans to West African farmers (6), social media companies in prioritizing news stories (7, 8), schools in promoting and laying off teachers (9, 10), and universities in offering admission and scholarships to students from around the world (11). Yet most people are unaware of the use and consequences of these algorithms, even though their lives have probably been impacted by them.

Most prediction algorithms are based on data accumulated in developed countries, developed by programmers and scientists who have been educated in developed countries, and most importantly, are often informed by socio-cultural norms for human behavior in developed countries (12). In a series of audits, researchers found that 96% of datasets in the behavioral sciences and 80% in education originate from a single country–the United States (13–15). This raises an important question: how well do models of human behavior developed in the U.S. generalize to other people, especially in less developed countries? No empirical evidence has directly answered this question so far, though recent studies have found global variation in moral preferences (16) and responses to poverty alleviation interventions (17, 18), suggesting that models of human behaviors differ across the world. However, many behavioral scientists and technology companies continue to operate under the implicit assumption that U.S.-based models generalize to other parts of the world (14, 19). Here we test whether this assumption is warranted or not.

Investigating the generalizability of U.S.-based models of human behavior to less developed countries offers three benefits to the general public and the scientific community.

First, if models are found to be generalizable, they can be used to guide policy interventions in data-scarce regions of the world, such as aiding policymakers in developing countries to effectively allocate constrained resources to target at-risk communities and individuals (20). Second, if models do not generalize, it alerts behavioral scientists, international governments and organizations, and technology companies to reevaluate their use of U.S.-based models of human behavior and devise more effective strategies to make predictions abroad. Thirdly, quantifying the degree to which U.S.-based models are generalizable can advance our scientific understanding of the variability of human behavior across geographic and cultural contexts, and improve the performance of predictive models going forward.

To achieve these goals, we begin by defining the generalizability of an empirical model developed in a given social context in terms of its prediction accuracy in other social contexts. For instance, the accuracy of a model of academic achievement trained on U.S. student data and applied to Canadian, Chinese, or Indonesian students provides a measure of this U.S.-based model's generalizability for these countries. We focus on the generalizability of models of academic achievement in this study for two reasons. On an individual level, academic achievement has significant consequences for employment opportunities (22), household income (22), personal well-being (23), and health (24). On a societal level, academic achievement is also referred to as "knowledge capital" and represents a critical antecedent for national development (25).

We draw on a large, nationally representative dataset of students in 65 countries ($N =$ 459,842) collected by the Programme for International Student Assessment (PISA). PISA is a triennial international survey that partners with governments to document and compare rates of student achievement. It is an influential data source for policymakers: 80% of national representatives from 37 different countries reported PISA to have led to changes in national education practices (21). We use a bottom-up machine learning approach with over 200 features to account for students' family, classroom, and school environment, their personal activities and psychological state. Compared to prior research that examined selected factors in isolation, this analytic approach estimates the importance of all features for academic achievement simultaneously and with minimal researcher degrees of freedom (26). These efforts promise to reveal what matters for student achievement, such as the number of weekly lessons, the number of books at home, the amount of parental emotional support, or perhaps how passionate the student feels about learning.

To evaluate the global generalizability of insights from the United States, we developed a predictive model of U.S. student achievement and tested its performance in other countries. PISA collected 165 student, family, and school variables from a nationally representative sample of U.S. students ($N = 5,712$). Students took a 90-minute standardized test on natural science concepts in Biology, Chemistry, and Physics. We trained a random forest model to identify students who rank in the top decile of science achievement in the U.S.—a common task in admissions processes (26)—and evaluated the cross-validated model on a holdout dataset.

**Results**

We found that the U.S.-trained model correctly identifies top students in the U.S. with 71.6% accuracy. This high level of accuracy is expected from a model that is trained and tested in the same population. Next, we tested how well the U.S.-trained model generalizes to less developed countries that may not have local data collection to build such a model and therefore rely on foreign models to inform education policy.

We tested the U.S. model in all 65 countries that participated in the 2015 PISA and collected identical measures as the U.S. As expected, the U.S. model was generally less accurate when applied in other countries: the average accuracy was 60.2% and varied between

74% in Sweden and 50% (i.e. no better than chance) in Indonesia. To evaluate how the U.S. model of academic achievement performs in less developed countries, we compared model performance to each country's United Nation's Human Development Index (HDI), a national index of economic, health, and education indicators (28). We found that the model's accuracy drops linearly in less and less developed countries [$r = 0.86$, $P < 0.001$, $n = 65$] (Fig. 1). In the least developed countries in our dataset (Indonesia, Vietnam, Moldova), U.S. model accuracy dropped to no better than chance, while in the most developed countries, model accuracy was as high as in the U.S.

This result is robust to changes in the classification threshold (top 10%, bottom 10%, top 50%) and evaluation criterion (AUC, F1, sensitivity, specificity, precision), presented in the *SI Appendix*. Multiple country-level indicators such as GDP per capita, the Gini index of income inequality, and cultural dimensions could explain the global variation in U.S. model accuracy, but HDI was the strongest predictor (*SI Appendix* Table S1).

This finding demonstrates that empirical insights accumulated in the U.S. can lose their explanatory power in less developed countries. It may be the case that the developing world is inherently less predictable regardless of where the data is accumulated. For instance, a high propensity for power outages and unstable internet connections might interrupt learning. Such contextual randomness is known to limit the effectiveness of self-regulation interventions to benefit students in India and China (29). If it were inherently harder to predict academic achievement, then even a model trained on local data should yield a low accuracy. We therefore compare the U.S. model to a local model as a reference point for how accurate a model can be if the same data is available. The previous pre-processing, factor selection, and model training steps were replicated in each country with the same test dataset (see SI for details). We found that local models achieve significantly higher accuracy than the U.S. model [paired $t = 12.8$, $P < 0.001$, $n = 65$], and that local model accuracy is uncorrelated with the Human Development

Index [$r = -0.01$, $P = 0.97$, $n = 65$] (Fig. 1). This rules out the possibility that the U.S. model's accuracy is lower simply because the developing world is less predictable. To the contrary, developing countries could produce models of achievement that are as accurate as those in the U.S, if local and representative data were available.

Nationally representative datasets as comprehensive as PISA are rare and expensive to collect in less developed countries (14, 16). This has left critical questions about why foreign insights fail to generalize unanswered. We use PISA data to probe the reasons why the U.S. model fails to predict student achievement in less developed countries. By comparing the models trained in each country, we can begin to understand the relative importance of various achievement factors and examine patterns along the dimension of national development.

To systematically examine which achievement factors are subject to global variation, we use machine learning to identify factors that vary with national development and those that do not. For each of the 53 countries where the PISA dataset includes the same 208 variables, we trained a ridge regression model with cross-validation to predict student achievement and inspect all model coefficients for the normalized predictors (see SI for details). Each coefficient represents the importance of one achievement predictor in a given country while accounting for the 207 remaining predictors.

We examined how the importance of achievement factors in local models varies by national development in terms of the Human Development Index and identified four distinct patterns, each illustrated with three examples in Fig. 2. Several factors strongly predict high academic achievement in the U.S., such as personal aspiration and enjoyment of science, but matter less in the developing world (Fig 2. top row). Some other factors strongly predict lower academic achievement in the U.S., such as fear of tests and parental occupation status, but are insignificant in the developing world (second row). Conversely, there are achievement factors that matter more in less developed countries, such as are whether students eat dinner and whether they have access to computers to learn from scientific simulations (third row). Finally, some achievement factors do not show systematic variation with national development, such as the number of lessons a student attends in school per week, whether the student watches TV after school, and whether she holds scientific beliefs.

The ground-up analysis of achievement factors revealed significant global variation in the importance of several factors that are established in the education literature. Specifically, we found that three theory-driven predictors of academic achievement—personal aspiration, books at home, and the weekly amount of class time (30–32)—are less important for achievement in the developing world, accounting for covariates. We further examined the pattern of global variation for the 208 achievement factors by categorizing them into four types: 33 objective student factors (e.g., grade, gender), 87 subjective student factors (e.g., enjoyment, interest), 37 school factors (e.g., class duration, disciplinary climate), and 46 family factors (e.g., family wealth, parental education). Global variation is measured by the correlation between factor importance in each country and the Human Development Index (HDI) of that country, as depicted in Fig.2, such that a large absolute correlation value indicates strong variability of a factor with national development level, whereas a value close to zero indicates invariance with national development level. We found all four types of achievement factors to exhibit global variation, but among them, students' subjective factors, such as enjoyment and interest, exhibited the widest range of variation. On average, students' subjective factors' importance correlates with HDI at $|r| = 0.264$, a significant difference from 0, $P < 0.001$. By contrast, objective student factors ($|r| = 0.163$), school factors ($|r| = 0.155$), and family factors ($|r| = 0.168$) had less variant importance across the globe. This means that while subjective factors (such as personal aspiration, enjoyment, and interest) are strong predictors of academic achievement in the developed world, they lack predicting power in the developing world. Indeed, international development agencies have recognized alternative factors that boost achievement. For instance, in an audit of mission statements of African's 30 largest cash transfer programs, a common poverty alleviation strategy, researchers found that over half of them (56%) tapped into community-based goals and processes(19). Our findings further illustrate the need to better understand and appreciate how achievement factors vary across the globe to develop computational models that have the potential to effectively allocate resources in less developed regions of the world.

**Discussion**

Big data and algorithms are affecting lives around the world. Yet global inequalities in the availability of data raise questions about the generalizability of analytic insights around the world. We show that a U.S.-trained model of academic achievement perform no better than chance in some parts of the developing world. This is because the factors that motivate human behavior and performance vary substantially across the globe, and thus what matters in the United States is different to what matters in the developing world. Unlike prior work, our findings are less susceptible to researcher bias because we used a bottom-up machine learning approach with nationally representative datasets from 53 countries.

We document a socio-technical dilemma: data and models of human behavior accumulated in developed countries lose their predictive power in the developing world, where they could have the greatest impact on improving the quality of lives. This pattern cannot be attributed to "inherent" randomness or unpredictability of outcomes in the developing world. Instead, we demonstrate that if equivalent datasets are available in developing counties, it is possible to build prediction models of human behaviors that are as accurate as those in the developed world.

Moreover, we uncover that among four types of predictors, the importance of students' subjective factors is most variable across the world. This finding contributes to theories in behavioral sciences, comparative education and cultural psychology. Students' personal aspirations and passion towards learning are theorized to matter more in developed countries because their cultural models and educational institutions prioritize self-directed action in learning (32). Students who express stronger interest, enjoyment, and self-confidence in learning earn higher grades in these countries (32, 33). In contrast, cultural models in developing countries cast the self as part of a social whole that is responsive to close others, strives to fulfill expectations, and attends to social norms. Students are motivated to earn good grades by a sense of duty, role models, and to please friends and family in these countries. Yet many large-scale educational datasets including PISA have limited information on interdependent sources of motivation. A notable exception in the PISA data is whether teachers emphasize the utility value of science—how it can benefit society—which is a stronger predictor of achievement in developing countries, but a relatively weaker predictor in more developed countries.

Although PISA records over 200 achievement factors in most countries, there probably are unobserved factors that matter in the developing world but have evaded measurement. Our results highlight the value of identifying additional factors to guide policies. The models we developed are correlational: they summarize patterns in the past to guide actions in the future. Randomized trials are needed to establish the strength of causal relationships between factors uncovered in international comparative studies.

Prior work has shown that a model for detecting emotional states that is trained on urban students fails to accurately detect emotions for rural and suburban students in the U.S. (34). The current research establishes a global boundary condition for whether human behavior prediction models developed in one population can generalize to another and identifies sources of deviation. It underscores the empirical relevance as well as the social significance of contextual variation in efforts to distribute the benefits of new innovations fairly among different populations (17).

Predictive algorithms originating in developed countries offer significant benefits to people and organizations, but they also pose risks. On the one hand, they synthesize patterns in human behavior that can be harnessed to effectively inform policies and improve people's lives in all parts of the world. On the other hand, if applied too broadly, they risk further disadvantaging people in resource-constrained environments by overlooking contextual variation and local knowledge. In the words of the anthropologist Anna Lowenhaupt Tsing, "scalability requires that project elements be oblivious to the indeterminacies of encounter; that's how they allow smooth expansion. Thus, too, scalability banishes meaningful diversity, that is, the diversity that might change things." Our findings highlight how global diversity changes things and we recommend that scientists, policy-makers, and technology companies scrutinize the contextual relevance of data sources and analytic models before scaling up their projects and policies globally.

**Materials and Methods**

All results and analyses presented are based on the PISA 2015 dataset. The dataset is publicly available at https://www.oecd.org/pisa/data/2015database/. Data pre-processing and analysis scripts are made available online at https://osf.io/3dv8w/. These codes enable replication of the study's findings.

We took 7 steps to train a model to predict academic achievement in the United States: defining the outcome measure, splitting the data into a training and a test set, selecting the set of features, preprocessing the feature set, tuning model parameters with ten-fold cross-validation, and building a random forest classifier. Details for these steps are available in the *SI Appendix.*

We evaluate the performance of the U.S.-trained model in 65 countries around the world ($N = 459,842$ students). We replicate the preprocessing procedure applied in the U.S.-trained model in each country's dataset with two modifications that simulate the realistic scenario of adopting a U.S.-trained model abroad. First, we keep constant the set of features to guarantee that the U.S.-trained model can be used directly for prediction abroad. Second, to handle missing values, we use the U.S. sample medians from the training set for imputation in all countries. This applies the assumption that the test set (foreign country) has the same distribution as the training set (U.S.). This assumption will be violated in many less developed countries, but it is necessary for a real-word application where distributional statistics for each country are unlikely to be known. To maximize ecological validity, we assume zero knowledge of the distributional properties of the features and outcomes in the test sets.

We evaluate model performance with established metrics based on the confusion matrix of predicted class and true class. Our primary evaluation metric is balanced accuracy, an average of sensitivity and specificity. This metric is robust to unbalanced classification problems. In our case, if the model simply predicted "0" for all students, the standard accuracy metric would be 0.90, while balanced accuracy would be 0.50. We additionally considered multiple standard evaluation metrics for binary classification problems: the F1 score (which we maximized during cross-validation), sensitivity, specificity, precision, and area under the curve (AUC) for tree-based models. *SI Appendix* Figure S1 demonstrates the robustness of the main result (Fig. 1) across these alternative evaluation metrics.

We conducted several robustness checks for our finding that the U.S.-trained model is decreasingly accurate in less developed countries (see *SI Appendix* Table S1). First, the result holds for different prediction goals beyond identifying students in the top decile; the pattern is the same for models trained to identify the top half and bottom decile in the country. Second, the pattern holds for five alternative prediction evaluation metrics beyond balanced accuracy: F1, sensitivity, precision, specificity, and AUC. Third, prediction performance correlates most strongly with HDI, and it also correlates with closely related indicators of national development and culture (GDP per capita, Gini index of inequality, individualism), providing evidence for convergent validity. However, it does not correlate with national education indicators defined within each country, providing evidence for divergent validity (see *SI Appendix* Table S1).

**Acknowledgements**

We thank Geoffrey L. Cohen and Hazel R. Markus for constructive feedback. We thank Yitian Ni and Zheng Zeng for their research assistantship in data cleaning. The first author was supported by a Stanford Graduate Fellowship and a Stanford Computational Social Science Fellowship.

**Figures**

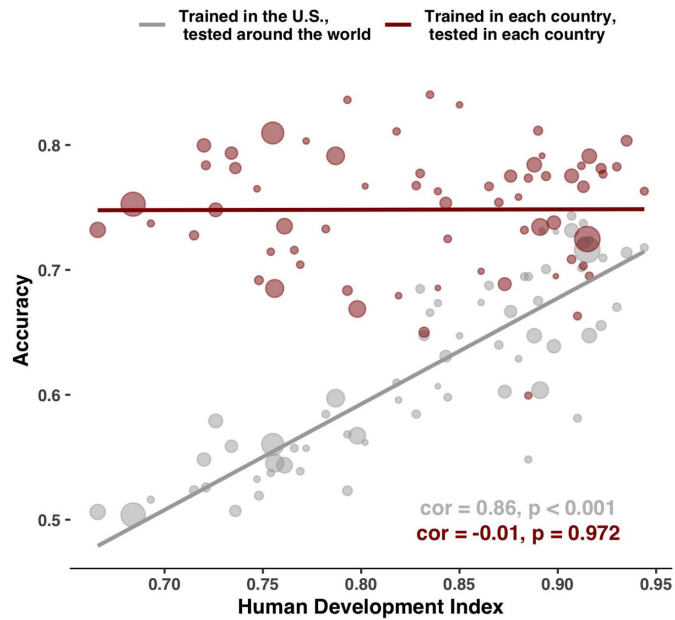

**Fig. 1. Academic achievement prediction accuracy by national development.** The accuracy of a U.S.-built model (grey) and locally-built models (red) are shown for each country's Human Development Index. Point size represents country population size.

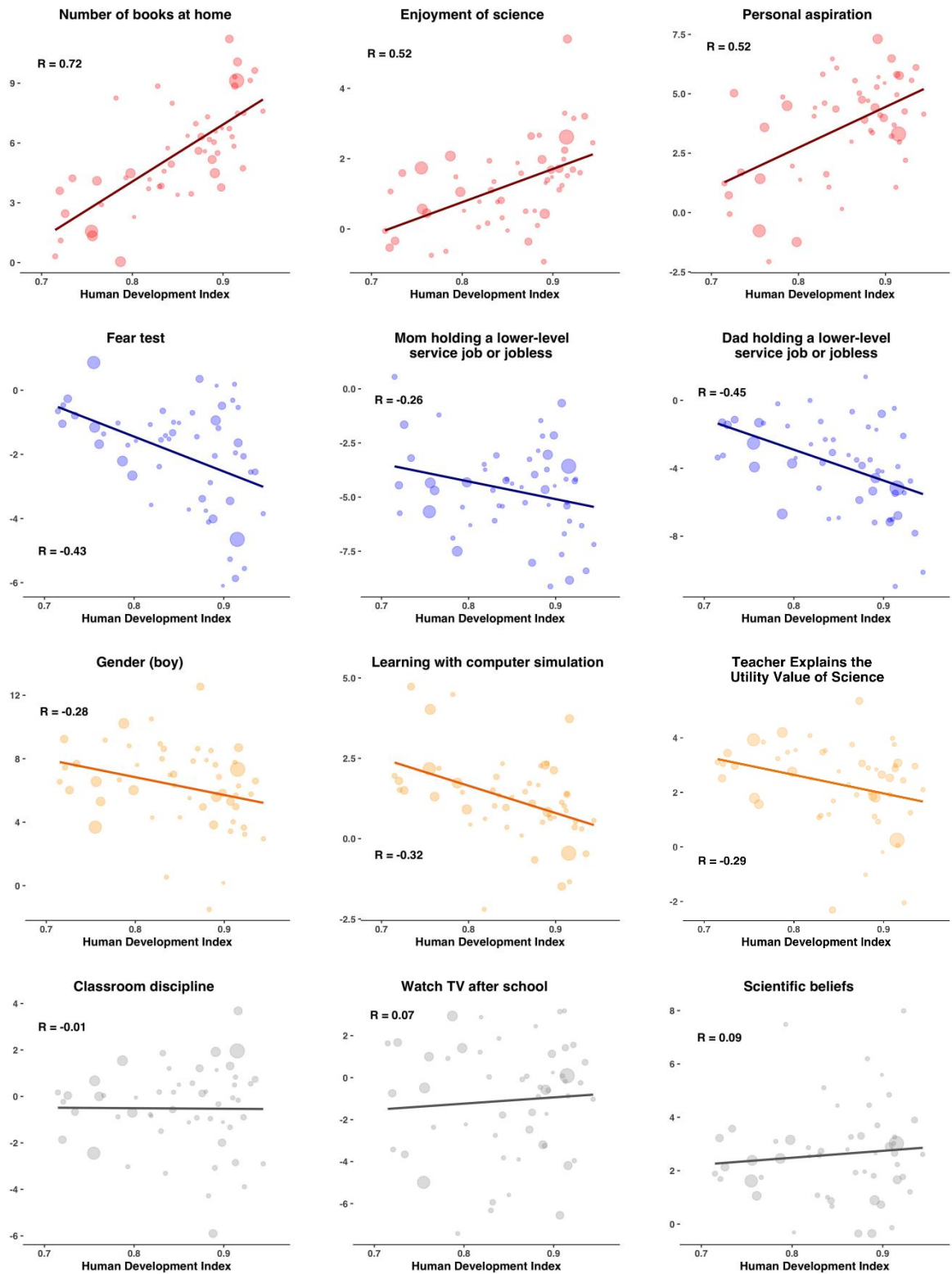

**Fig. 2. Four patterns of global variation in factor importance across 53 countries.** Factors that predicted significantly stronger (red) and weaker (blue) achievement in developed countries but are weak predictors in developing countries. Factors that had stable coefficients

across the globe (grey). Factors that predicted significantly stronger achievement in developing countries but are weak predictors in developed countries (yellow).

## Part 1. Training a model to predict academic achievement in the United States

*Defining the outcome measure.* The PISA dataset provides students' test scores on a 90-minute standardized test in science. In fact, it contains not just one test score but ten PVs (Plausible Values) to gauge students' academic achievement. Its official data analysis guide states that using the first among the ten PVs ("PV 1") is sufficient for assessing academic achievement (1). To fit a classifier model of high-performing students in each country, we needed to construct a binary outcome measure. We therefore use PV 1 in Science and constructed a binary indicator variable of whether a student is in the top 10% (coded as 1) versus bottom 90% (coded as 0) in each country.

*Splitting the data into a training and a test set.* All data analysis was conducted using the open-source statistical computing language R (https://www.r-project.org/). We used the *createDataPartition* function in the *caret* package (2) to perform a stratified sampling to obtain a random subset of the 80% of the data points from the United States ($N_{United States}$ = 5,712) to be our training set ($N_{train}$ = 4,572). The remaining 20% of the data points serve as our test set and were not used in the training phase.

*Selecting the set of features.* Variables with missing values for all observations or a constant value for all observations were excluded because they provide no information. Variables identifying the school or country were excluded. Variables coded inconsistently across countries were excluded: for example, for the variable "PROGN", a value of 1 indicates grades 7-9 and 2 indicates grades 10-12 in the United States, while in Uruguay, values 1-5 indicate lower secondary grades and values 6-10 upper secondary grades. A total of 283 features remain after this step.

*Preprocessing the feature set.* For each variable, we computed the percentage of missing values, which ranges from 0 to 37.6% (mean = 5.5%; SD = 4.9%). We imputed missing values with the median of the available values for each variable. No standardization procedure was applied in preprocessing in all of our studies because either the prediction method is robust to monotonic transformations or the software implementation applies a standardization by default. There was no evidence of multiple collinearities in the feature set and no individual feature was highly correlated with the outcome measure (Pearson correlations between -0.33 and 0.40).

*Tuning model parameters with ten-fold cross-validation.* We used random forest classification in the training dataset with ten-fold cross-validation to select a value for the maximum depth of trees that maximizes the F1 score—the widely used indicator of predictive accuracy. Importantly, we needed to use both case weights and class weights in this process. Case weights are needed due to PISA's use of stratified sampling in data collection. To guarantee the national representativeness of samples, student observations are weighted based on the non-response rates provided in the PISA dataset (W_FSTUWT). Class weights are needed as our prediction problem (identifying the top 10%) is unbalanced by definition. Setting class weights to 9:1—the reciprocal ratio of class sizes, returned better results than omitting class weights. We used the R package *ranger* for fitting models because it accepts both case weights and class weights (3).

*Building a random forest classifier.* We trained a random forest classifier over the entire training set, setting the number of trees to 500, the number of variables to split at each node to the square root of the number of features, the splitting rule objective to reducing the Gini coefficient, and the maximum depth to 7, which was determined using cross-validation. In the main text, we refer to this model as the U.S.-trained model, which encompasses this full set of operations, from feature selection, preprocessing, to cross-validation and final model training.

Part 2. Testing the U.S.-trained model in other countries

We evaluate the performance of the U.S.-trained model in 65 countries around the world (N = 459,842 students). We replicate the preprocessing procedure detailed in Part 1 in each country's dataset with two modifications that simulate the realistic scenario of adopting a U.S.-trained model abroad. First, we keep constant the set of features to guarantee that the U.S.-trained model can be used directly for prediction abroad. Second, to handle missing values, we use the U.S. sample medians from the training set for imputation in all countries. This applies the assumption that the test set (foreign country) has the same distribution as the training set (U.S.). This assumption will be violated in many less developed countries, but it is necessary for a real-word application where distributional statistics for each country are unlikely to be known. To maximize ecological validity, we assume zero knowledge of the distributional properties of the features and outcomes in the test sets.

We evaluate model performance with established metrics based on the confusion matrix of predicted class and true class. Our primary evaluation metric, reported in the main text, is balanced accuracy, an average of sensitivity and specificity. This metric is robust to unbalanced classification problems. In our case, if the model simply predicted "0" for all students, the standard accuracy metric would be 0.90, while balanced accuracy would be 0.50. We additionally considered multiple standard evaluation metrics for binary classification problems: the F1 score (which we maximized during cross-validation), sensitivity, specificity, precision, and area under the curve (AUC) for tree-based models. Figure S1 demonstrates the robustness of the main result (Fig. 1 in the main text) across these alternative evaluation metrics.

Part 3. Robustness checks

We conducted several robustness checks for our finding that the U.S.-trained model is decreasingly accurate in less developed countries (Table S1). First, the result holds for different prediction goals beyond identifying students in the top decile; the pattern is the same for models trained to identify the top half and bottom decile in the country. Second, the pattern holds for five alternative prediction evaluation metrics beyond balanced accuracy: F1, sensitivity, precision, specificity, and AUC. Third, prediction performance correlates most strongly with HDI, and it also correlates with closely related indicators of national development and culture (GDP per capita, Gini index of inequality, individualism), providing evidence for convergent validity. However, it does not correlate with national education indicators defined within each country, providing evidence for divergent validity (see Table S1).

Part 4. Estimating global variation in the importance of achievement factors

To quantify which features have weights that vary systematically across the globe, and which have stable weights across countries, we need to estimate each feature's importance for predicting achievement in each country and compare those estimates across countries.

We used ridge regression for two reasons. First, it is a machine learning method that allows us to quantify feature importance in the prediction of academic achievement in each country. Second, it allows the outcome variable to be continuous so that we can test whether the previously discussed pattern extends beyond binary outcomes. To enable fair comparisons between countries, we limit the number of countries to those with a set of identical features: instead of maximizing available features for each country, as we did with the random forest classifier, we applied a standardized procedure to include only identical features (p = 208) among the majority of countries (N = 53).

We gauged the prediction performance of the ridge model by $R^2$. It can be interpreted as the percentage of variance explained by the model, or the square of the correlation between predicted and true test scores. Results are robust for different evaluation criteria including Root Relative Squared Error (RRSE) and Relative Absolute

Error (RAE). Methods for building this model and testing it across the globe are presented below.

*Sample selection*. To compare model performance and feature importance across countries, we sought to select a large sample of countries with a common set of features. The number of features measured in each country varied substantially across the world, which presented a trade-off between retaining more countries (but fewer features) and retaining more features (but fewer countries). Figure S2 visualizes this tradeoff in terms of the number of common features and the respective number of countries. To ensure a sufficient amount of variation in HDI, we prioritize the number of countries in the sample and selected the largest number of countries before the feature set size declines steeply. This preserved 208 features in 53 out of the 65 countries. Furthermore, five features that were coded inconsistently across countries (PROGN, LANGN, COBN_F, COBN_M, and COBN_S) were excluded, leaving 203 features. Having determined the sample and feature set, we moved on to training predictive models of achievement.

*Defining the outcome measure*. As in the previous section, "PV 1 in Science" is used as the measure of students' academic achievement.

*Splitting the data into a training and a test set*. As in the previous section, we used the *createDataPartition* function to perform stratified sampling and obtain a random subset of 80% of the data points from the United States ($N_{United States}$ = 5,712) to serve as the training set ($N_{train}$ = 4,572). The remaining 20% of the data points serve as the test set and remained untouched in the training phase.

*Creating a new feature to account for missing values*. In addition to the 203 features, we created a new student-level feature that measures what percentage of the 203 features are missing. We expect missingness to be informative of individual factors like school attendance. The update feature set comprises 204 features, 54 categorical and 150 continuous variables.

*Preprocessing the feature set*. We checked the rate of missing values for each feature. The average percentage of missing values was 5.5% (SD = 4.5%, range from 0 to 29.5%). Different imputation methods were applied to categorical and continuous variables. For categorical variables, missing values were first imputed as -99 to indicate that it is missing and use this information in the prediction. Then, binary indicators (dummy variables) were created for each level of each categorical feature using the *dummyVars* function from the *caret* package (4). For continuous variables, we used this k-nearest neighbor (KNN) imputation with k = 5. The "knnImpute" method in the *preProcess* function identifies for each student who has a missing value, their five closest neighbors in the training set with complete observations. Then KNN algorithm imputes the missing value with the mean of the neighbors' values. The KNN procedure standardizes all numerical features to have a mean of 0 and a standard deviation of 1 to compute the Euclidean distance between observations. This procedure is slower than the imputation of median values, but it is more accurate because it accounts for correlations between features.

*Selecting the set of features*. We removed variables with zero variance because they provide no information for prediction. We did not filter out highly correlated variables since the L2-regularization in the Ridge regression automatically addresses issues of multi-collinearity (5). We checked the correlation between the outcome and each of the features. No individual feature was highly correlated with the outcome measure; Pearson correlation coefficients ranged between -0.33 and 0.40.

*Tuning model parameters with cross-validation.* We trained a ridge regression model using the *cv.glmnet* function (setting alpha = 0 for ridge regression) and using the PISA case weights (W_FSTUWT) with ten-fold cross-validation. The *cv.glmnet* function automates the fitting and cross-validation of the ridge model, optimizing its performance by lowering the weighted RMSE. In the main text, this ridge model is also referred to as the U.S.-trained model.

Part 5. Testing the U.S.-trained model in the U.S. test set

We applied the U.S.-trained model to the U.S. test set ($N_{test}$ = 1,140) that was set aside. We relied on information from the training set to process the test set because the distribution of the test set was assumed to be unknown; by splitting the data randomly, the test set is representative of the training set. First, the feature selection and creation process for the test set was the same as for the training set. Second, we standardized continuous features using the mean and standard deviation in the training set and imputed missing values for the test set: imputing missing values with the average of the five nearest neighbors in the training set and creating dummies for categorical features. Finally, we excluded zero-variance columns.

For each student, the U.S.-trained model predicts the academic achievement outcome. We use three performance measures to evaluate if our findings are robust across different evaluation criteria or constrained to some of them. We measured this U.S.-trained model's performance on the test set by RRSE (Root Relative Squared Error; RMSE [Root Mean Squared Error] of our prediction divided by the RMSE of always predicting the median value), RAE (Relative Absolute Error; MAE [Mean Absolute Error] of our prediction divided by the MAE of always predicting the median value), and R-squared (coefficient of determination). The two relative performance measures were used to adjust for variation in academic achievement across countries. The U.S.-trained model performed well on the test set, RMSE = 60.53 (RRSE = 0.63), MAE = 47.84 (RAE = 0.60), and $R^2$ = 0.608. It is important to note that the model explains 60.8% of variation in the student achievement, which is a higher explanatory power than many models in the social sciences (6).

To validate the ridge regression model, we compare its performance with an OLS regression model. We fitted the OLS model by selecting up to 20 predictors using the method of forward-selection (adding one variable at a time which offers the greatest increase in in-sample $R^2$). The OLS model achieves a substantially lower prediction performance with RMSE = 68.70 (RRSE = 0.71), MAE = 54.59 (RAE = 0.69), and $R^2$ = 0.497. This provides further evidence that the U.S.-trained ridge model is effective at predicting student achievement in the United States.

Part 6. Testing the U.S.-trained model in other countries

Without any further modification of the U.S.-trained model, we applied it to every country in our selected sample. The data for each country was pre-processed following the same procedure as the U.S. test set: standardization of numerical features, imputation of missing values, and creation of dummies for categorical features. To preserve ecological validity during standardization and imputation, we again assumed no knowledge of the test set and use the distributional information from the U.S.-trained model. The features in the test set the and the training set were kept the same, as the sample selection process ensured that the test set would not contain any missing features. However, in cases where categorical variables contain levels that are different from the training set, we either added a column of zeros (if a level present in the U.S. was missing in another country) or dropped the category dummy indicator (if another country had a level that did not occur in the U.S.). Finally, we applied the U.S.-trained model to the pre-processed data from other countries and measured the U.S.-model's prediction performance. Results are presented in the main text and below.

Part 7. Training and testing predictive models for each country

After having tested the U.S.-trained model in other countries, we independently trained and tested a ridge regression model for each country. The models are referred to as nationally-trained models. For each country, student observations were randomly split into an 80% training set and a 20% test set. Then we followed the same procedure as for the U.S.-trained model. Using only the training sets, we completed procedures of standardization, imputation, and dummy creation. Finally, a weighted ridge model was fitted on the training set and tested on its corresponding test set. We measured model prediction performance using $R^2$, RRSE and RAE. Figure S3 shows the differences in performance between the nationally-trained models and the U.S.-trained model across countries.

*Estimating global variability in achievement factors.* To assess how feature importance varies across countries by developmental level, we compute the Pearson correlation coefficient between the Human Development Index and the ridge regression coefficient from the nationally-trained models for each feature. All features that appear in more than 30 countries are considered. Overall, the correlation coefficients between HDI and local feature importance are normally distributed (M = -0.01, SD = 0.24; see Figure S4). Nearly a third of the local achievement factors varied across the globe with statistical significance ($P < 0.05$, n = 53), including 25 that were more important in the developing world and 71 that were more important in the developed world. This variability provides a novel window into the geography of academic achievement and provides cautionary evidence for any effort to export and scale educational interventions outside of the United States.

*Feature categorization.* The first and second author independently coded 203 variables into four categories (indices of intercoder reliability: percentage agreement = 0.946; Cohen's kappa = 0.941): students' objective factors, students' subjective factors, school factors, and family factors (for a complete list of features by category, see Table S2). Disagreements between coders were resolved by a process of deliberation between them. Given the correlations between ridge regression coefficients and HDI, we find that features related to personal preference and behavior exhibit higher variability in correlation coefficients compared to the other categories of features (Figure S5).

To formally test for differences in the variability between feature categories, we applied a bootstrap approach. For each bootstrap replicate, we generated a random sample by sampling with replacement from the original sample of correlation coefficients. Each bootstrap sample has the same size as the original sample. From each bootstrap sample, we can calculate the parameter of interest. Different values of the parameter in all the replicates constitute an empirical distribution on which we can perform statistical tests. The parameters of interest in this case are the standard deviation and interquartile range (IQR) of each feature group. We generated 200 bootstrap samples and performed a t-test to check if the differences in variances and IQRs were significant. A test was performed to compare the second group (students' subjective factors) to each of the other three groups, yielding three results for each statistic. This provided confirmatory evidence that the variance and IQR of the second group were significantly larger than any of the other three groups [all p-values < 0.001, bootstrap n = 200], see Table S3.

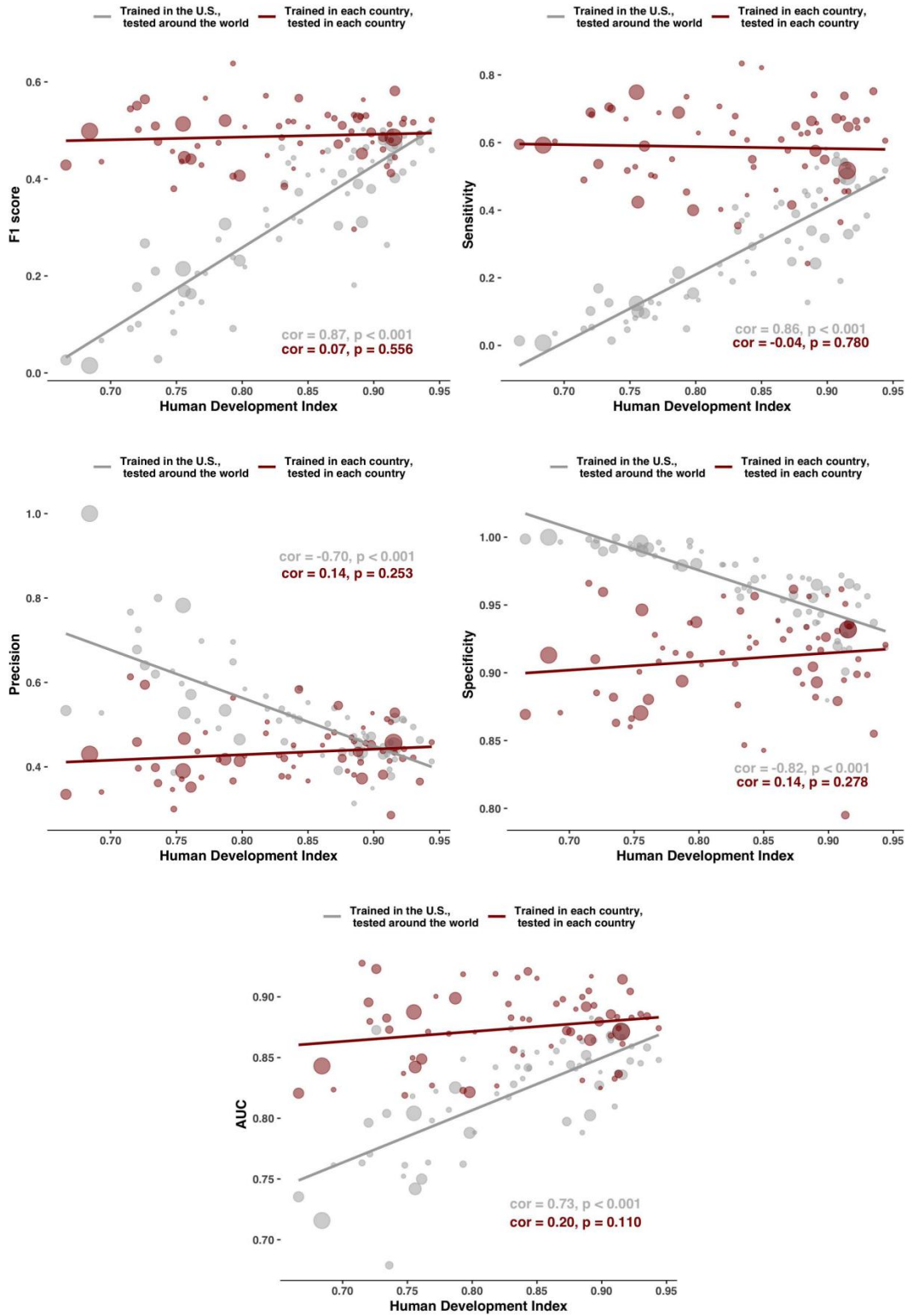

**Fig. S1.** Comparison of evaluation metrics for random forest classifier performance. Grey lines indicate results of the U.S.-trained model; red lines indicate nationally-trained models. Differences between slopes in each plot are significant [P < 0.001, n = 65].

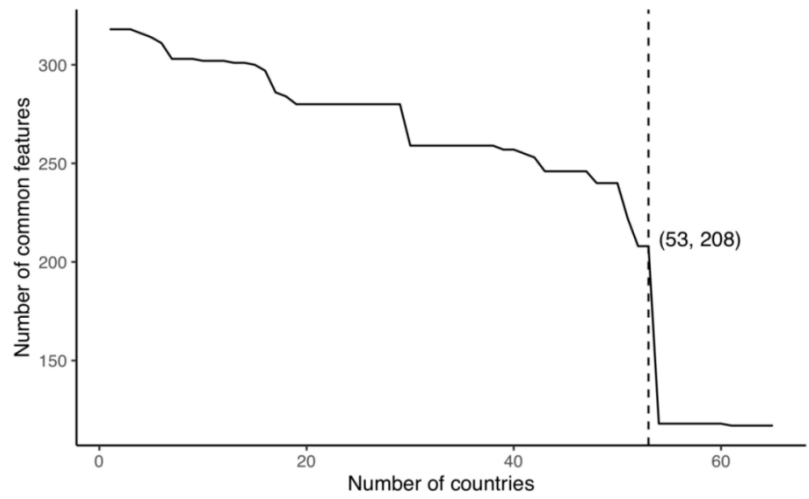

**Fig. S2.** Selecting common features across countries to preserve as many countries as possible while maximizing the number of features available.

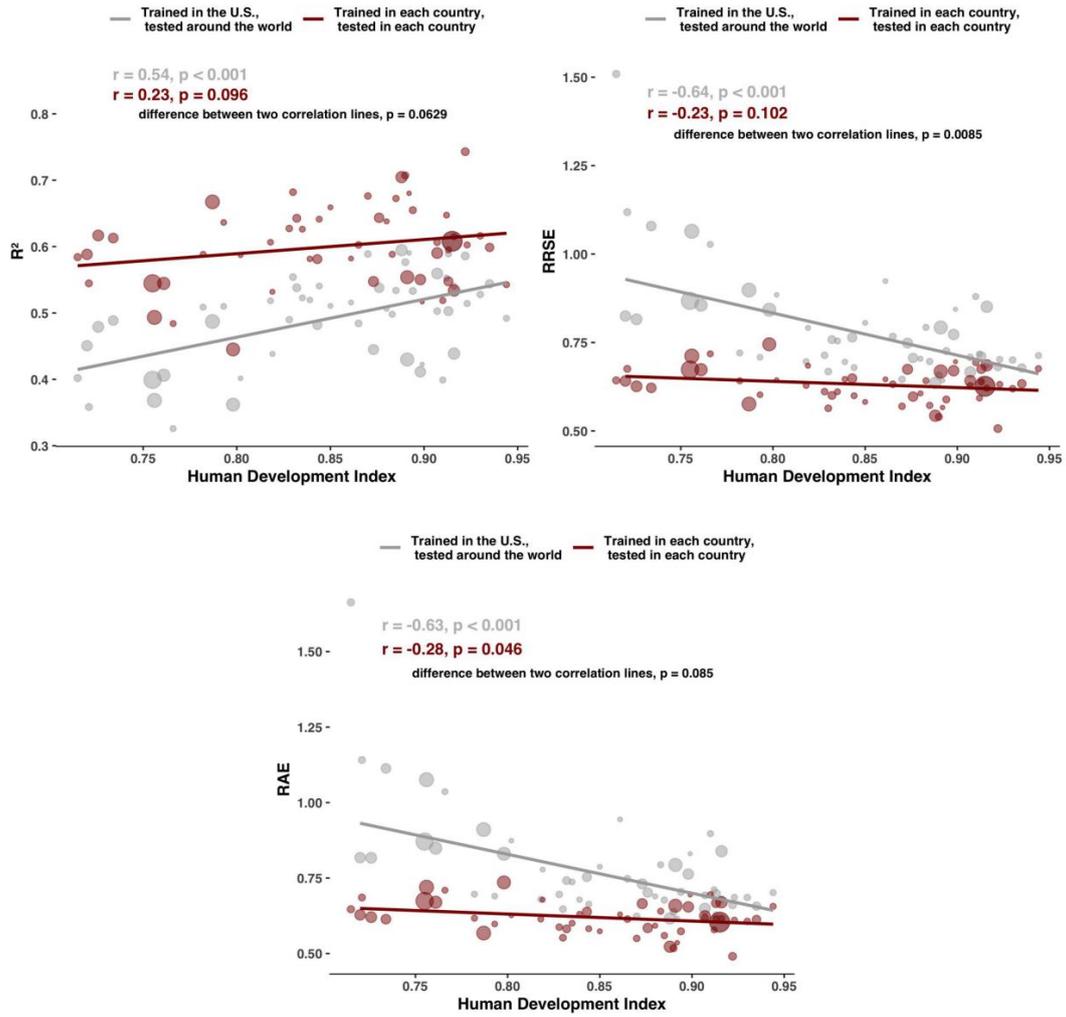

**Fig. S3.** Comparison of evaluation metrics for ridge regression performance. Grey lines indicate results of the U.S.-trained model; red lines indicate nationally-trained models. P-values for differences between slopes in each plot are shown in the plot.

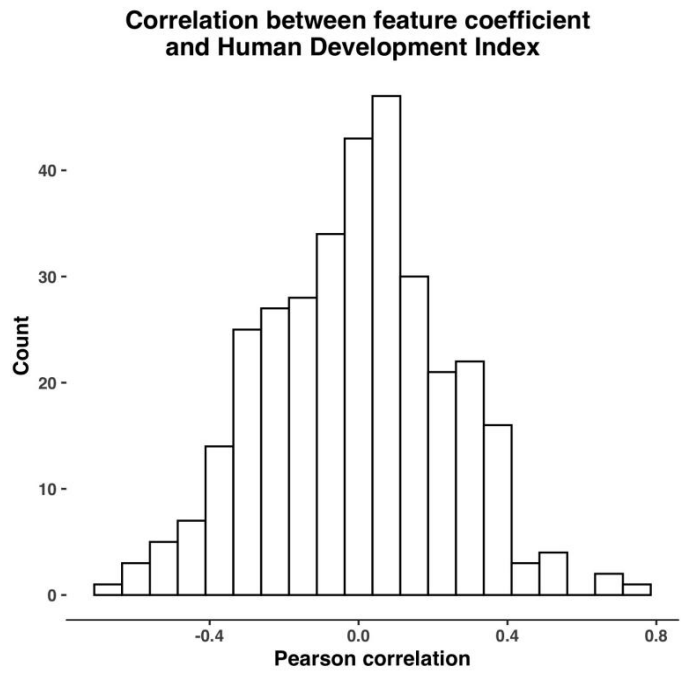

**Fig. S4.** Distribution of Pearson correlation coefficients (n = 203) between countries'

Human Development Index and national-model regression coefficients.

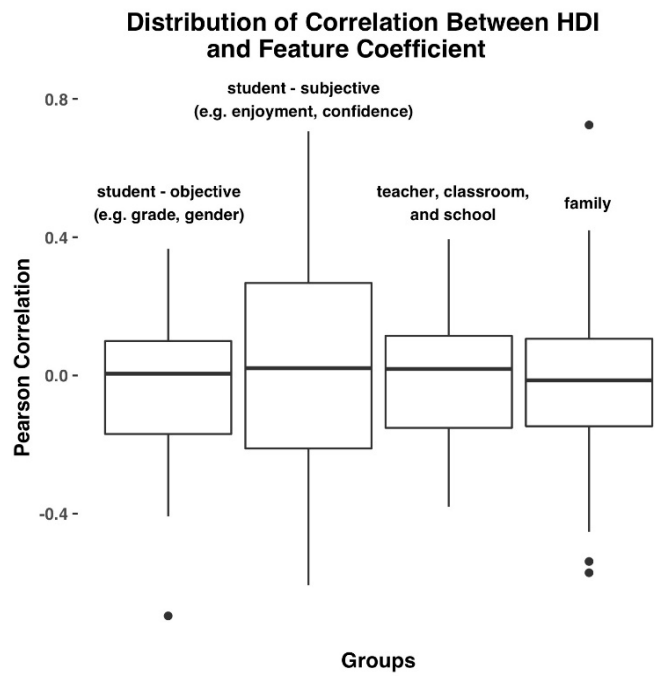

**Fig. S5.** Distribution of correlation coefficients between HDI and national-model regression coefficients for features grouped into four categories (n = 33, 87, 37, 46). The second category (students' subjective factors) exhibits higher variability than the other three.

**Table S1.** Pearson correlations between U.S.-trained model performance (comparing six evaluation metrics for two prediction goals) across 65 countries and six country-level development and cultural indicators. HDI is the best predictor of model performance across all prediction goals and evaluation metrics.

| | Random Forest Classifier for top 50% | | | | | | Random Forest Classifier for bottom 10% | | | | | |
|---|---|---|---|---|---|---|---|---|---|---|---|---|
| | Balanced accuracy | F1 | Sensitivity | Precision | Specificity | AUC | Balanced accuracy | F1 | Sensitivity | Precision | Specificity | AUC |
| HDI | 0.74*** | 0.77*** | 0.85*** | -0.66*** | -0.79*** | 0.67*** | 0.28** | 0.73*** | -0.51*** | 0.76*** | 0.71*** | 0.67*** |
| GDP p.c. | 0.57*** | 0.58*** | 0.64*** | -0.45*** | -0.58*** | 0.55*** | 0.22* | 0.50*** | -0.31** | 0.52*** | 0.47*** | 0.47*** |
| Individualism | 0.51*** | 0.51*** | 0.55*** | -0.37*** | -0.43*** | 0.45*** | 0.22* | 0.47*** | -0.24* | 0.40*** | 0.44*** | 0.43*** |
| Educational expenditure (% GDP) | -0.28 | -0.27 | -0.23 | 0.16 | 0.12 | -0.28 | -0.15 | -0.27 | 0.23 | -0.22 | -0.33 | -0.21 |
| Gini index | -0.42*** | -0.40*** | -0.42*** | 0.30** | 0.32** | -0.36** | -0.25* | -0.51*** | 0.21 | -0.48*** | -0.47*** | -0.34** |
| Secondary education enrollment (% population) | 0.069 | 0.085 | 0.050 | -0.026 | -0.013 | 0.079 | 0.11 | 0.16 | -0.11 | 0.15 | 0.17 | 0.19 |

Significance of correlation: *$P$ < 0.1; **$P$ < 0.05; ***$P$ < 0.01. All n = 65.

**Table S2.** Categorization of features in the dataset into four groups.

| 1. Students' objective factors |
| --- |
| Student International Grade (Derived) |
| Student (Standardized) Gender |
| Country of Birth International - Self |
| How old were you when you arrived in <country of test>? |
| How old were you when you started <ISCED 1>? Years |
| Before going to school did you: Eat breakfast |
| Before going to school did you: Study for school or homework |
| Before going to school did you: Watch TV\<DVD>\Video |
| Before going to school did you: Read a book\newspaper\magazine |
| Before going to school did you: Internet\Chat\Social networks (e.g. <Facebook>, <country-specific social network>) |
| Before going to school did you: Play video-games |
| Before going to school did you: Meet friends or talk to friends on the phone |
| Before going to school did you: Talk to your parents |
| Before going to school did you: Work in the household or take care |
| Before going to school did you: Work for pay |
| Before going to school did you: Exercise or practice a sport |

After leaving school did you: Eat dinner

After leaving school did you: Study\school\hmk

After leaving school did you: Watch TV\<DVD>\Video

After leaving school did you: Read a book\newspaper\magazine

After leaving school did you: Internet\Chat\Social net (e.g. <Facebook>)

After leaving school did you: Play video-games

After leaving school did you: Meet friends or talk to friends on the phone

After leaving school did you: Talk to your parents

After leaving school did you: Work in the household or take care of other family members

After leaving school did you: Work for pay

After leaving school did you: Exercise or practice a sport

Age

ISCED level

ISCED designation

ISCED orientation

Grade compared to modal grade in the country

Index Immigration status

2. Students' subjective factors

To what extent do you disagree or agree about yourself? I often worry that it will be difficult for me taking a test.

To what extent do you disagree or agree about yourself? I worry that I will get poor <grades> at school.

To what extent do you disagree or agree about yourself? Even if I am well-prepared for a test I feel very anxious.

I get very tense when I study for a test.

I get nervous when I don't know how to solve a task at school.

I want top grades in most or all of my courses.

I want to be able to select from among the best opportunities available when I graduate.

I want to be the best, whatever I do.

I see myself as an ambitious person.

I want to be one of the best students in my class.

<NAME 1> is motivated? Gives up easily when confronted with a problem and is often not prepared

<NAME 2> is motivated? Mostly remains interested in the tasks she starts and sometimes does more than expected

<NAME 3> is motivated? Wants to get top grades at school and continues working on tasks until perfect.

To what extent do you disagree or agree about yourself? I prefer working as part of a team to working alone.

To what extent do you disagree or agree about yourself? I am a good listener.

To what extent do you disagree or agree about yourself? I enjoy seeing my classmates be successful.

To what extent do you disagree or agree about yourself? I take into account what others are interested in.

To what extent do you disagree or agree about yourself? I find that teams make better decisions than individuals.

To what extent do you disagree or agree about yourself? I enjoy considering different perspectives.

To what extent do you disagree or agree about yourself? I find that teamwork raises my own efficiency.

To what extent do you disagree or agree about yourself? I enjoy cooperating with peers.

In the last two full weeks of school, how often: I <skipped> a whole school day

In the last two full weeks of school, how often: I <skipped> some classes

In the last two full weeks of school, how often: I arrived late for school

How informed are you about this environmental issue? The increase of greenhouse gases in the atmosphere

How informed are you about this environmental issue? The use of genetically modified organisms (<GMO>)

How informed are you about this environmental issue? Nuclear waste

How informed are you about this environmental issue? The consequences of clearing forests\other land use

How informed are you about this environmental issue? Extinction of plants and animals

How informed are you about this environmental issue? Water shortage

This issue will improve or get worse over next 20 years? Air pollution (student)

This issue will improve or get worse over next 20 years? Extinction of plants and animals (student)

This issue will improve or get worse over next 20 years? Clearing of forests for other land use (student)

This issue will improve or get worse over next 20 years? Water shortages (student)

This issue will improve or get worse over next 20 years? Nuclear waste (student)

This issue will improve or get worse over next 20 years? The increase of greenhouse gases in the atmosphere

This issue will improve or get worse over next 20 years? The use of genetically modified organisms (<GMO>)

Disagree or agree with the statements? I have fun when I am learning <broad science>

Disagree or agree with the statements? I like reading about <broad science> topics.

Disagree or agree with the statements? I am happy working on <broad science> topics.

Disagree or agree with the statements? I enjoy acquiring new knowledge in <broad science>.

Disagree or agree with the statements? I am interested in learning about <broad science>.

To what extent are you interested in: Biosphere (e.g. ecosystem services, sustainability)

To what extent are you interested in: Motion and forces (e.g. velocity, friction, magnetic and gravi forces)

To what extent are you interested in: Energy and its transformation (e.g. conservation, chemical reactions)

To what extent are you interested in: <broad science> topics? The Universe and its history

To what extent are you interested in: <broad science> topics? How science can help us prevent disease

Making an effort in my <school science> subject(s) is worth it because this will help me in the work I want to do lat

What I learn in my <school science> subject(s) is important for me because I need this for what I want to do later on

Studying my <school science> subject(s) is worthhile for me because what I learn will improve my career prospects.

Many things I learn in my <school science> subject(s) will help me to get a job.

Recognise the science question that underlies a newspaper report on a health issue.

Explain why earthquakes occur more frequently in some areas than in others.

Describe the role of antibiotics in the treatment of disease.

Identify the science question associated with the disposal of garbage.

Predict how changes to an environment will affect the survival of certain species.

Interpret the scientific information provided on the labelling of food items.

Discuss how new evidence can lead you to change your understanding about the possibility of life on Mars.

Identify the better of two explanations for the formation of acid rain.

A good way to know if something is true is to do an experiment.

How much do you disagree or agree with the statements below? Ideas in <broad science> sometimes change.

Good answers are based on evidence from many different experiments.

It is good to try experiments more than once to make sure of your findings.

Sometimes <broad science> scientists change their minds about what is true

The ideas in <broad science> science books sometimes change.

How often do you do these things? Watch TV programmes about <broad science>

How often do you do these things? Borrow or buy books on <broad science> topics

How often do you do these things? Visit web sites about <broad science> topics

How often do you do these things? Read <broad science> magazines or science articles in newspapers

How often do you do these things? Attend a <science club>

How often do you do these things? Simulate natural phenomena in computer programs\virtual labs

How often do you do these things? Simulate technical processes in computer programs\virtual labs

How often do you do these things? Visit web sites of ecology organisations

How often do you do these things? Follow news via blogs and microblogging

Environmental optimism (WLE)

Enjoyment of science (WLE)

Interest in broad science topics (WLE)

Instrumental motivation (WLE)

Science self-efficacy (WLE)

Epistemological beliefs (WLE)

Index science activities (WLE)

Students' expected occupational status (SEI)

Personality: Test Anxiety (WLE)

Student Atttidudes, Preferences and Self-related beliefs: Achieving motivation (WLE)

Collaboration and teamwork dispositions: Enjoy cooperation (WLE)

| Collaboration and teamwork dispositions: Value cooperation (WLE) |
| --- |
| ISCO-08 Occupation code - Self |

3. School factors

Number of <class periods> required per week in <test language>

Number of <class periods> required per week in mathematics

Number of <class periods> required per week in <science>

In a normal, full week at school, how many <class periods> are you required to attend in total?

How many minutes, on average, are there in a <class period>?

This school year, approximately how many hours per week do you spend learning in addition? <School Science>

This school year, approximately how many hours per week do you spend learning in addition? Mathematics

This school year, approximately how many hours per week do you spend learning in addition? <Test language>

This school year, approximately how many hours per week do you spend learning in addition? <Foreign language>

This school year, approximately how many hours per week do you spend learning in addition? Other

How often does this happen in your <school science> lessons? Students don't listen to what the teacher says.

How often does this happen in your <school science> lessons? There is noise and disorder.

How often does this happen in your <school science> lessons? The teacher waits long for students to quiet down.

How often does this happen in your <school science> lessons? Students cannot work well.

How often does this happen in your <school science> lessons? Students don't start working for a long time after.

When learning <school science>? Students are given opportunities to explain their ideas.

When learning <school science>? Students spend time in the laboratory doing practical experiments.

When learning <school science>? Students are required to argue about science questions.

When learning <school science>? Students are asked to draw conclusions from an experiment they have conducted.

When learning <school science>? The teacher explains <school science> idea can be applied

When learning <school science>? Students are allowed to design their own experiments.

When learning <school science>? There is a class debate about investigations.

When learning <school science>? The teacher clearly explains relevance <broad science> concepts to our lives.

When learning <school science>? Students are asked to do an investigation to test ideas.

How often does this happen in your <school science> lessons? The teacher shows interest every students learning.

How often does this happen in your <school science> lessons? The teacher gives extra help.

How often does this happen in your <school science> lessons? The teacher helps students with their learning.

How often does this happen in your <school science>? The teacher continues teaching\students understand.

How often does this happen in your <school science>? Teacher gives an opportunity to express opinions.

Disciplinary climate in science classes (WLE)

Teacher support in a science classes of students choice (WLE)

Inquiry-based science teaching an learning practices (WLE)

Out-of-School Study Time per week (Sum)

Learning time (minutes per week) - <Mathematics>

Learning time (minutes per week) - <test language>

Learning time (minutes per week) - <science>

| |
|---|
| Learning time (minutes per week) - in total |
| 4. Family factors |

What is the <highest level of schooling> completed by your mother?

Does your mother have this qualification? <ISCED level 6>

Does your mother have this qualification? <ISCED level 5A>

What is the <highest level of schooling> completed by your father?

Does your father have this qualification? <ISCED level 6>

Does your father have this qualification? <ISCED level 5A>

In your home: A desk to study at

In your home: A quiet place to study

In your home: A computer you can use for school work

In your home: Educational software

In your home: A link to the Internet

In your home: Classic literature (e.g. <Shakespeare>)

In your home: Books of poetry

In your home: Works of art (e.g. paintings)

In your home: Books to help with your school work

In your home: <Technical reference books>

In your home: A dictionary

In your home: Books on art, music, or design

In your home: <Country-specific wealth item 1>

In your home: <Country-specific wealth item 2>

In your home: <Country-specific wealth item 3>

How many in your home: Televisions

How many in your home: Cars

How many in your home: <Cell phones> with Internet access (e.g. smartphones)

How many in your home: Computers (desktop computer, portable laptop, or notebook)

How many in your home: <Tablet computers> (e.g. <iPad®>, <BlackBerry® PlayBookTM>)

How many in your home: E-book readers (e.g. <KindleTM>, <Kobo>, <Bookeen>)

How many in your home: Musical instruments (e.g. guitar, piano)

How many books are there in your home?

Country of Birth International - Mother

Country of Birth International - Father

International Language at Home

Mother's Education (ISCED)

Father's Education (ISCED)

Highest Education of parents (ISCED)

ISEI of mother

ISEI of father

Index highest parental occupational status

Index highest parental education in years of schooling

ISCO-08 Occupation code – Mother

ISCO-08 Occupation code – Father

Home educational resources (WLE)

Home possessions (WLE)

ICT Resources (WLE)

Family wealth (WLE)

Index of economic, social and cultural status (WLE)

**Table S3.** Group difference in measures of spread from bootstrap estimates.

| Comparing pairs | Difference in Standard Deviation | Difference in Interquartile Range |
|---|---|---|
| Students' subjective factors vs. students' objective factors | 0.106*** | 0.203*** |
| Students' subjective factors vs. school factors | 0.115*** | 0.221*** |
| Students' subjective factors vs. family factors | 0.094*** | 0.211*** |

Significance: ***$P < 0.01$; Bootstrap $n = 200$.

**SI References**